# A Novel Adaptive Fine-Tuning Algorithm for Multimodal Models: Self-Optimizing Classification and Selection of High-Quality Datasets in Remote Sensing

Yi Ren, Tianyi Zhang, Zhixiong Han, Weibin Li, Zhiyang Wang, Wenbo Ji, Chenhao Qin, Chenbin Liang, Licheng Jiao, *Fellow, IEEE*

*Abstract*—General multimodal large models pre-trained with large-scale data can effectively perform domain-specific downstream tasks in the computer vision field, but they cannot be seamlessly scalable to remote sensing due to the modal disparity. However, multimodal large models tailored to remote sensing are still under-explored, and the main challenge is how to obtain optimal performance without substantially increasing computational overheads. Therefore, we propose an adaptive fine-tuning algorithm for multimodal large models. The core steps of this algorithm involve two stages of truncation. First, the vast amount of data is projected into a semantic vector space, and the MiniBatchKMeans algorithm is used for automated clustering. This classification ensures that the data within each cluster exhibit high semantic similarity. Next, we process the data in each cluster, calculating the translational difference between the original and perturbed data in the multimodal large model's vector space. This difference serves as a generalization metric for the data. Based on this metric, we select the data with high generalization potential for training. We applied this algorithm to train the InternLM-XComposer2-VL-7B model on two 3090 GPUs using one-third of the GeoChat multimodal remote sensing dataset. The results demonstrate that our algorithm outperforms the state-of-the-art baselines. various baselines. The model trained on our optimally chosen one-third dataset, based on experimental validation, exhibited only 1% reduction in performance across various remote sensing metrics compared to the model trained on the full dataset. This approach significantly preserved general-purpose capabilities while reducing training time by 68.2%. Furthermore, the model achieved scores of 89.86 and 77.19 on the UCMerced and AID evaluation datasets, respectively, surpassing the GeoChat dataset by 5.43 and 5.16 points. It only showed a 0.91-point average decrease on the LRBEN evaluation dataset. Our code is open-sourced and available at (https://github.com/renllll/).

*Index Terms*—Multimodal large models, High-quality multimodal datasets, Adaptive fine-tuning algorithm

Yi Ren, Chenbin Liang are with the Lab. of AI, Hangzhou Institute of Technology of Xidian University, Hangzhou, 311231, China. Tianyi Zhang, Wenbo Ji, Weibin Li, Chenhao Qin, Licheng Jiao are with the School of Artificial Intelligence, Xidian University, Xi'an, 710071, China. Zhixiong Han is with the Key Laboratory of Coal Resources Exploration and Comprehensive Utilization, Ministry of Natural Resources, Xi'an, 710018, China. Zhiyang Wang is with the Shaanxi Water Development Group Co., Ltd, Xi'an, 710018, China. (email: weibinli@xidian.edu.cn;)

## I. INTRODUCTION

The emergence of large language models (LLMs) has brought significant advancements to the field of artificial intelligence, demonstrating remarkable capabilities across various natural language processing tasks. For instance, models like ChatGPT[1] and GPT-4[2] exhibit strong zero-shot and few-shot[3] learning abilities, which allow them to generalize well across many domains. However, when applied to specialized fields such as healthcare, law, and hydrology, these general-purpose models often experience performance degradation, since their insufficient training in domain-specific knowledge results in a lack of understanding of tasks within these specialized areas..

To address this issue, researchers have begun exploring specialized training and fine-tuning of LLMs for specific domains, and notable achievements have been made. For example, in the medical field[4-s], Google and DeepMind introduced Med-PaLM[5], a model designed for medical dialogue, which excels in tasks such as medical question answering, diagnostic advice, and patient education. Han et al. proposed MedAlpaca[6], a model fine-tuned on a large corpus of medical data based on Stanford Alpaca[7], aimed at serving medical question answering and consultation scenarios. Wang et al. developed BenTsao[8], which was fine-tuned using Chinese synthetic data generated from medical knowledge graphs and literature, providing accurate Chinese medical consultation services. In the legal field, Zhou et al. introduced LaWGPT[9], which was developed through secondary pre-training and instruction fine-tuning on large-scale Chinese legal corpora, enabling robust legal question answering capabilities. In the field of hydrology, Ren et al. proposed WaterGPT[10], a model based on Qwen-7B-Chat[11] and Qwen2-7B-Chat[12], which successfully achieved knowledge-based question answering and intelligent tool invocation within the hydrology domain through extensive secondary pre-training and instruction fine-tuning on domain-specific data.

With the success of LLMs in various fields, researchers have gradually started to explore the development of domain-specific multimodal models. For instance, in the medical field, Wang et al. introduced XrayGLM[13] to address challenges in interpreting various medical images. Li et al. proposed LLaVA-Med[14], aiming to build a large language and vision model with GPT-4 level capabilities in the biomedical domain.

In the field of remote sensing, real-world tasks often require multi-faceted comprehensive analysis to achieve effective solutions. Therefore, practical applications typically necessitate multi-task collaboration for accurate judgment. Despite significant advancements in deep learning[15,16] within the remote sensing field, most current research still focuses on addressing single tasks and designing architectures for individual tasks[17], which limits the comprehensive processing of remote sensing images[18,19]. Consequently, multi-modal large models may exhibit exceptional performance in the remote sensing domain.

In the field of remote sensing, significant progress has also been made by researchers. For example, Liu et al. introduced RemoteCLIP[20], the first vision-language foundation model specifically designed for remote sensing, aimed at learning robust visual features with rich semantics and generating aligned textual embeddings for various downstream tasks. Zhang et al. proposed a novel framework for domain-specific pre-training of vision-language models, DVLM[21], and trained the GeoRSCLIP model for remote sensing. They also created a paired image-text dataset called RS5M for this purpose. Hu et al. released a high-quality remote sensing image caption dataset, RSICap[22], to promote the development of large vision-language models in the remote sensing domain, and provided the RSIEval benchmark dataset for comprehensive evaluation of these models' performance. Kuckreja et al. introduced GeoChat[23], a multimodal model specifically designed for remote sensing, capable of handling various remote sensing images and performing visual question answering and scene classification tasks. They also proposed the RS multimodal instruction following dataset, which includes 318k multimodal instructions, and the geo-bench evaluation dataset for assessing the performance of multimodal models in remote sensing.Zhang et al. proposed EarthGPT[24], which seamlessly integrates multi-sensor image understanding and various remote sensing visual tasks within a single framework. EarthGPT can comprehend optical, synthetic aperture radar (SAR), and infrared images under natural language instructions, and accomplish a range of tasks including remote sensing scene classification, image description, visual question answering, object description, visual localization, and object detection.Liu et al. introduced the Change-Agent platform[25], which integrates a multi-level change interpretation model (MCI) and a large language model (LLM) to provide comprehensive and interactive remote sensing change analysis, achieving state-of-the-art performance in change detection and description while offering a new pathway for intelligent remote sensing applications.

However, most current research focuses on direct training using large multimodal datasets, leading to significant computational resource consumption. Studies have shown that fine-tuning on a small amount of high-quality data can achieve good results. For instance, Wei et al. demonstrated that after fine-tuning InstructionGPT-4[26] on 6% of selected data, its performance surpassed the original MiniGPT-4 across various tasks. Regarding the selection of high-quality fine-tuning datasets, Kung et al. proposed the Active Instruction Tuning method[27], proving that datasets with high prompt uncertainty possess stronger generalization abilities. Yang et al. proposed a Self-Distillation method[28] to mitigate the catastrophic forgetting phenomenon after LLM fine-tuning. Yu et al. introduced WaveCoder[29], which projects datasets into vector space and uses KCenterGreedy for clustering to select core datasets. Although many studies have explored how to select high-quality datasets, no algorithm has effectively filtered high-quality datasets suitable for fine-tuning multimodal models, allowing the model to significantly enhance domain-specific capabilities while retaining generalization abilities.

To address this gap, we propose a novel adaptive fine-tuning algorithm for multimodal large models, capable of automatically categorizing and filtering remote sensing multimodal instruction datasets to identify high-quality data for training from massive remote sensing datasets. The core steps of the algorithm include projecting the large-scale data into semantic vector space and using the MiniBatchKMeans algorithm for automated clustering. Each data cluster is then processed by introducing perturbation parameters to the original data and calculating the translational differences between the original and perturbed data in the multimodal model's vector space. This difference serves as a generalization performance metric, determining the quality of the dataset. Finally, through a layer of ranking, we select the batch of datasets with the highest generalization performance metrics for training.

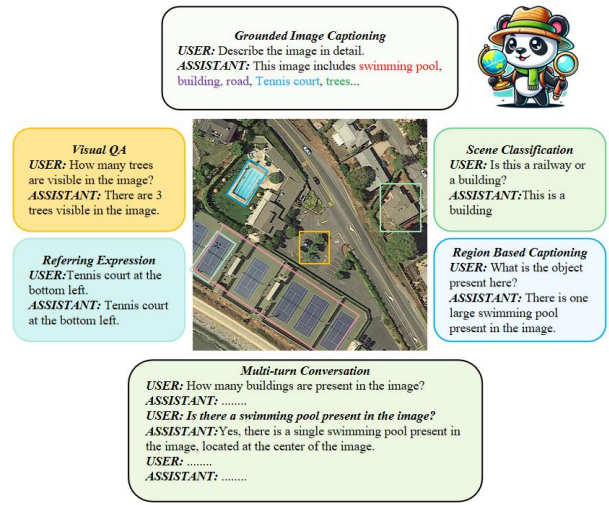

**Fig. 1.** Various tasks that our remote sensing multi-modal large model can complete

We utilize the RS multimodal instruction-following dataset proposed by GeoChat for training and adopt the Evaluation Benchmark from GeoChat along with MMBench_DEV_EN[30], MME[31], and SEEDBench_IMG[32] as evaluation datasets for domain-specific and general domains, respectively. Through comparisons with random selection, the WaveCoder algorithm, and our proposed algorithm on the GeoChat classification dataset, our results demonstrate that our algorithm outperforms other baseline methods, maximizing domain capability enhancement while preserving generalization ability. Additionally, our algorithm's selected one-third dataset reduces training time by approximately two-thirds compared to training on the entire dataset, with only a 1% average decrease in performance in the remote sensing domain, while significantly maintaining generalization capability. The multimodal large model we trained excels in various remote sensing image question-answering and comprehension tasks (Figure 1).

The main contributions of this paper are as follows:

1. We propose a new multimodal instruction fine-tuning dataset quality metric—generalization performance metric.

2. We introduce a novel algorithm that selects high-quality remote sensing multimodal fine-tuning datasets to achieve faster and more efficient training results.

3. By training on small datasets, we compare the effects of baseline algorithms and our algorithm in both general and remote sensing domains, validating that our algorithm achieves favorable results in the remote sensing domain.

## II. DATASET CREATION

### A. Training Data

The RS multimodal instruction following dataset is a multimodal instruction-following dataset designed for remote sensing image understanding. It integrates various tasks such as image description, visual question answering, and visual dialogue, aiming to enhance the model's ability to handle complex reasoning, object attribute understanding, and spatial relationships. The dataset contains a total of 318,000 instruction pairs.

### B. Evaluation Datasets

Our evaluation datasets include two parts: the remote sensing evaluation dataset and the general multimodal evaluation dataset.

(1) Remote Sensing Evaluation Datasets:

LRBEN (Land Use and Land Cover Remote Sensing Benchmark Dataset): This dataset is designed for land use and land cover classification tasks in remote sensing. It includes high-resolution images annotated for various types of land cover, such as urban areas, forests, water bodies, and agricultural fields. LRBEN is used to benchmark models' performance in visual question answering, scene classification, and other tasks in remote sensing.

UC Merced Land Use Dataset: This dataset contains aerial imagery of various land use classes, such as agricultural, residential, and commercial areas. The images are high-resolution and cover 21 different classes, each with 100 images, making it suitable for scene classification tasks. It is widely used for evaluating remote sensing models' ability to classify and understand different land use types.

AID (Aerial Image Dataset): AID is a large-scale dataset for aerial scene classification. It contains images from various scenes, such as industrial areas, residential areas, and transportation hubs. The dataset is designed to help in developing and benchmarking algorithms for scene classification, image retrieval, and other remote sensing tasks. AID includes a significant number of images for each category, providing a comprehensive benchmark for evaluating model performance.

C. General Multimodal Evaluation Datasets:

MMBench_DEV_EN: MMBench is a benchmark suite for evaluating the multimodal understanding capabilities of large vision-language models (LVLMs). It contains approximately 2974 multiple-choice questions covering 20 capability dimensions. Each question is single-choice, ensuring the reliability and reproducibility of the evaluation results. MMBench uses a strategy called cyclic evaluation to more reliably test the performance of vision-language models.

MME (Multi-Modal Evaluation): MME is a comprehensive evaluation benchmark for large multimodal language models, aiming to systematically develop a holistic evaluation process. The MME dataset includes up to 30 of the latest multimodal large language models and consists of 14 sub-tasks to test the models' perceptual and cognitive abilities. The MME data annotations are all manually designed to avoid potential data leakage issues that might arise from using public datasets.

SEEDBench_IMG: SEEDBench is an image dataset specifically designed for training and evaluating multimodal models. It contains high-quality image data with detailed annotations, suitable for various multimodal tasks such as image classification, object detection, and scene understanding. The SEEDBench dataset aims to assist researchers in developing and optimizing multimodal models by providing a comprehensive benchmark.

## III. METHODS

### A. Adaptive Self-Tuning for Multimodal Models

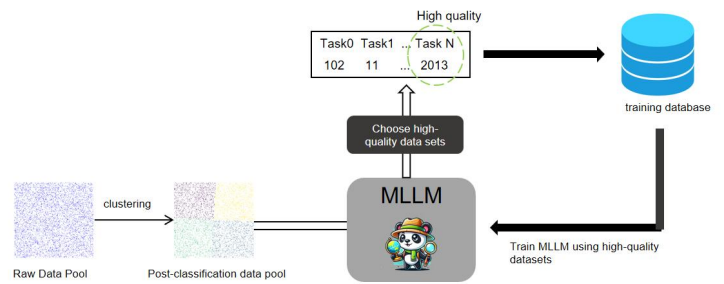

**Fig. 2.** Adaptive Self-Tuning for Multimodal Models algorithm flow

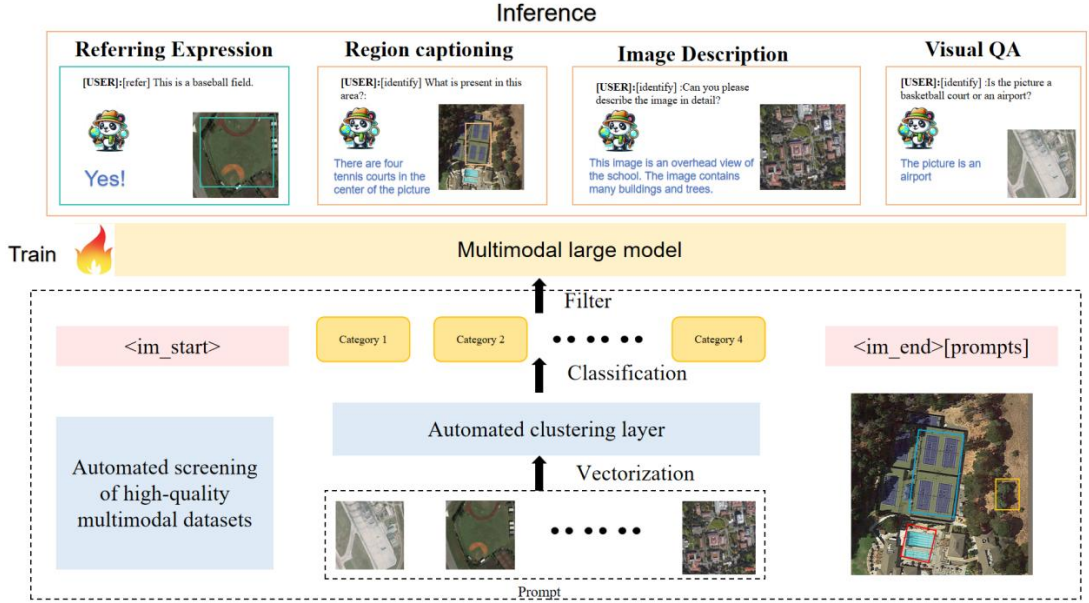

**Fig. 3.** Complete process of Adaptive Self-Tuning for Multimodal Models algorithm

In real-world scenarios, the volume of instruction fine-tuning data is often large and continually expanding, leading to increased training costs. Additionally, as the data volume grows, data conflicts also become more pronounced, often resulting in poorer training outcomes. To address this issue, we propose a new algorithm that enables large models to autonomously select data to better adapt to domain-specific tasks. The core of this algorithm is to allow the model to independently identify the most generalizable task instructions, achieving optimal performance with a minimal amount of training data. The flowchart of this process is shown in Figure 2. The complete training and inference process of our algorithm is illustrated in Figure 3.

*B. Selection of Generalizable Tasks*

The autonomous selection of task instruction datasets with greater generalization has been a research hotspot. For instance, Sid-dhant and Lipton's work on uncertainty-based active learning [33] provides significant insights.

Inspired by these studies, we propose a new generalization measure: vector space translation difference. Since large models predict the next word based on context, changes in the context vector affect subsequent content generation. We evaluate the uncertainty of instructions by randomly deleting words from the instruction context as perturbation information and observing the degree of change in the model's vector space. Generally, entries with stronger uncertainty yield better generalization effects after training. Specifically, the vector space translation difference measures the translation difference in the vector space of the model's projection vectors when given complete and perturbed task instructions, assessing the generalization of the instruction. This quantifies the model's responsiveness to uncertain instructions, enabling better evaluation of the model's generalization performance.

The detailed flowchart is shown in Figure 4, and the specific steps are as follows:

1. For the massive data pool X, we use the bge-large-en-v1.5[34] model to project each data entry into ector space, and then perform automated clustering using the MiniBatchKMeans algorithm. Specifically, we perform clustering calculations for different numbers of clusters using the MiniBatchKMeans algorithm, record the SSE (Sum of Squared Errors) and silhouette coefficient for each cluster number, and select the optimal number of clusters based on the highest silhouette coefficient. The data is eventually divided into p clusters. The specific steps are as follows:

（1）Data projection onto vector space:

$$V_i = BGE(X_i)$$

Here, $X_i$ represents the ith data item in the data pool, and $V_i$ represents the vector representation projected through the bge-large-en-v1.5 model.

（2）Calculation of the Sum of Squared Errors (SSE):

$$SSE = \sum_{j=1}^{p} \sum_{V_i \in C_j} \|V_i - \mu_j\|^2$$

Here, k represents the number of clusters, $C_j$ denotes the jth cluster, and $\mu_j$ is the centroid of the jth cluster. $V_i$ represents the vector belonging to the jth cluster. The SSE measures the sum of the distances between data points and their respective cluster centroids, serving as one of the indicators to evaluate clustering performance. A smaller SSE indicates that the points within a cluster are more tightly grouped. By plotting the SSE values for different numbers of clusters p, one can preliminarily assess the reasonable range for the number of clusters.

（3）Calculation of the Silhouette Coefficient:

$$s(i) = \frac{b(i) - a(i)}{\max(a(i), b(i))}$$

Here, a(i) represents the average distance from data point i to all other points within the same cluster, and b(i) represents the average distance from data point i to the nearest points in a different cluster. The silhouette coefficient S for the entire dataset is the average of the silhouette scores s(i) for all data points:

$$S = \frac{1}{n}\sum_{i=1}^{n} s(i)$$

Here, n represents the total number of data points.

（4）Selection of the optimal number of clusters:

$$p = \arg\max_{k} S(k)$$

Here, S(k) represents the silhouette coefficient for different numbers of clusters k, and p is the optimal number of clusters that maximizes S(k).

2. For the given p-th cluster and the K-th original instruction I0, add a perturbation parameter n (i.e., the number of words randomly deleted from each instruction). Generate N perturbed instructions randomly, denoted as $I_1$ to $I_N$.

3. Then, concatenate the input image $X_0$ and answer with $I_0$ to $I_N$ and project them into the vector space of the multimodal large model, as shown in the following formula:

$$E_1 = f(x_0, I_1) \ldots E_{N-1} = f(x_0, I_{N-1}), E_N = f(x_0, I_N)$$

4. For the instructions $I_0$ to $I_N$ and their corresponding images and answers, calculate the Euclidean distances between the projection vectors $E_0$ to $E_N$ and the perturbed vectors $E_1$ to $E_N$ sequentially, as follows:

$$\| E_1 - E_0 \|_2 \ldots \| E_{N-1} - E_0 \|_2, \| E_N - E_0 \|_2$$

5. Sum the Euclidean distances between the perturbed vectors $E_1$ to $E_N$ and $E_0$, then calculate the average value as the generalization measure, where n represents the perturbation parameter value, and K represents the K-th data entry.

$$S_{n,k} = \frac{1}{N}\sum_{i=1}^{N} \| E_i - E_0 \|_2$$

6. Finally, sort each instruction in the p-th cluster based on their generalization measures.

$$\text{Sort}(S_{1,k}, \ldots, S_{n,k})$$

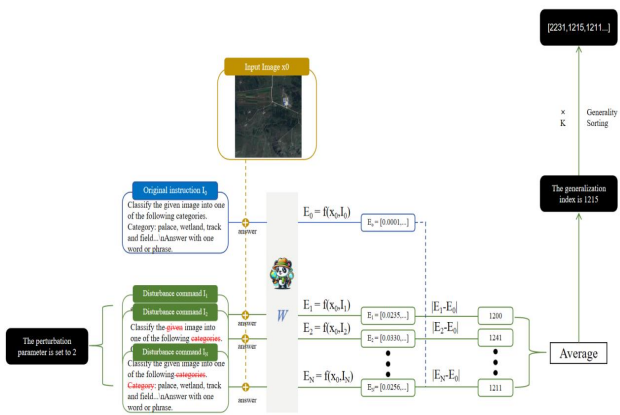

**Fig. 4.** Adaptive Self-Tuning for Multimodal Models Calculating Generalization Index Process

*C. Selection of optimal disturbance parameters*

To select the optimal disturbance parameter n, we observe the relative embedding differences when adding different disturbance parameters to determine the best value for n. The specific steps are as follows:

1. First, for the given K-th original instruction I0, sequentially add random parameters from 1 to n, resulting in disturbed instructions $I_1$ to $I_n$.

2. Then, concatenate the input image $X_0$ and the answer with $I_0$ to $I_n$ respectively, and project them into the vector space of the multimodal large model to obtain vectors $E_0$ to $E_n$. The formula is as follows:

3. For the obtained vectors $E_0$ to $E_n$, sequentially calculate the Euclidean distance between each perturbed vector $E_1$ to $E_n$ and the original vector $E_0$ to $E_n$. The formula is as follows:

$$\| E_1 - E_0 \|_2 \ldots \| E_{n-1} - E_0 \|_2, \| E_n - E_0 \|_2$$

4. Then, calculate the average embedding difference $S_{n,k}$ for the K entries under the disturbance parameter n. Sequentially calculate the relative embedding differences $D_{n,K}$ from 1 to n, and select the disturbance parameter with the maximum relative embedding difference as the optimal disturbance parameter. The formula is as follows, where K represents the p-th data pool containing K entries, and n represents the disturbance parameter:

$$S_{n,K} = \sum_{i=1}^{K} \| E_n^i - E_0^i \|_2$$

$$D_{n,k} = S_{n,K} - S_{n-1,K}$$

$$n = P(n \mid Max(D_{1,K}, \ldots D_{n,K}))$$

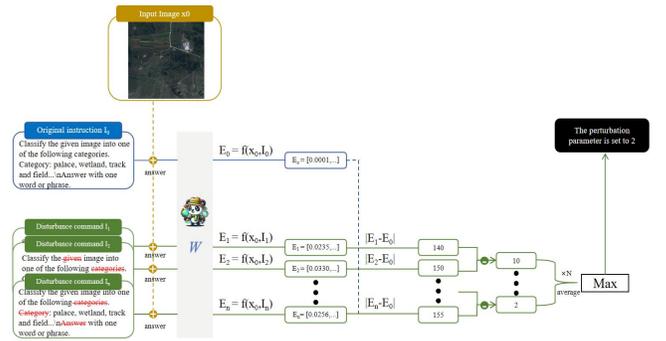

**Fig. 5.** Adaptive Self-Tuning for Multimodal Models algorithm selects the best disturbance parameter n process

*D. Compare algorithms*

Algorithm 1: Random Sampling

The random sampling method involves randomly selecting a subset of the dataset for training. This approach often captures the most diverse and broadly representative data from the dataset. Therefore, we use the random sampling algorithm as our baseline for comparison.

Algorithm 2: KCenterGreedy Clustering Algorithm

WaveCoder proposes a method for selecting a core dataset using the KCenterGreedy clustering algorithm. In this approach, we use the bge-visualized-m3[35] model to project each image-text pair into vector space, then apply the KCenterGreedy algorithm for clustering, and select a representative subset of the dataset.

## IV. EXPERIMENTS AND ANALYSIS

### A. Training Details

We performed LoRA[36] fine-tuning on the InternLM-XComposer2-VL-7B[37] model using the RS multimodal instruction following dataset. The fine-tuning parameters are as follows:

TABLE I
TRAIN PARAMETERS

| Hyper parameter | Value |
|---|---|
| Precision | fp16 |
| Epochs | 3 |
| Max length | 4096 |
| Batch size | 8 |
| Weight_decay | 0.1 |
| Warmup_ratio | 0.01 |

### B. Experiment on Disturbance Parameter Settings

To validate the effectiveness of our algorithm, we used a subset of clustered data focused on classification tasks, containing 3.2k entries, as the training set. We first evaluated the optimal disturbance parameter using our algorithm, and the relative vector embedding differences are shown in Figure 6.

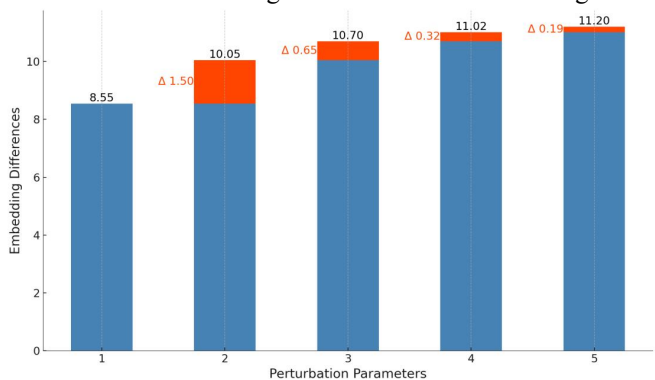

**Fig. 6.** Relative vector embedding difference under different disturbance parameters

As shown in the figure, the optimal disturbance parameter is 2, with the value gradually converging and the change magnitude decreasing, approaching zero after 4.

Therefore, we set the optimal disturbance parameter to 2. To further verify this, we used our algorithm to rank the generalizability of the training set with disturbance parameters from 1 to 4. We selected the top 5000 entries with the highest generalizability for training and evaluated the performance on the UC Merced and AID datasets. The results are shown in Figure 7.

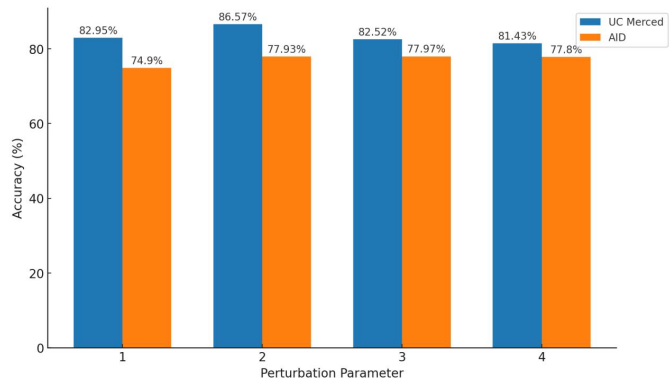

**Fig. 7.** Model training effect under different disturbance parameters

From the figure, it is evident that the model achieves the best training performance when the disturbance parameter is set to 2, reaching an accuracy of 86.57% on the UC Merced dataset, which is 4 points higher than when the disturbance parameter is 1 or 3. On the AID dataset, it also achieved 77.93%, only 0.04 points lower than when the disturbance parameter is 3. Overall, the model achieves optimal training performance when the disturbance parameter is set to 2.

### C. Comparison of Algorithm Performance

To further validate the effectiveness of our algorithm, we compared random sampling, the KCenterGreedy clustering algorithm, and our algorithm. We selected 5000 data entries for training in each case and compared the model's performance on the UC Merced and AID datasets. The results are shown in Table 2.

TABLE II
COMPARISON OF TRAINING EFFECTS OF DIFFERENT ALGORITHM MODELS UNDER 5000 PIECES OF DATA

| Method | AID | UC Merced | Avg. |
|---|---|---|---|
| Baseline(random) | 77.43 | 85.90 | 81.67 |
| KCenterGreedy | 78.07 ↑ 0.64 | 82.00 ↓ 3.90 | 80.04 ↓ 1.63 |
| Ours | 77.93 ↑ 0.50 | 86.57 ↑ 0.67 | 82.25 ↑ 0.58 |

TABLE III
COMPARISON OF TRAINING EFFECTS OF DIFFERENT ALGORITHM MODELS UNDER DIFFERENT SCALES OF DATA

| Method | Size | AID | UC Merced | Avg. |
|---|---|---|---|---|
| Baseline (random) | 10k | 78.10 | 87.52 | 82.81 |
| Ours | 10k | 78.73 ↑ 0.63 | 89.29 ↑ 1.77 | 84.04 ↑ 1.20 |
| Direct | 32k | 81.37 ↑ 3.27 | 90.71 ↑ 3.19 | 86.04 ↑ 3.23 |

TABLE IV
COMPARISON OF GENERAL PERFORMANCE OF DIFFERENT ALGORITHM MODELS UNDER DIFFERENT SCALES OF DATA

| Method | Model | Size | MMBench | Seedbench | MME |
|---|---|---|---|---|---|
| / | InternLM-XComposer2-VL-7B | / | 83.97 | 75.9 | 2242.70 |
| Baseline (random) | InternLM-XComposer2-VL-7B | 10k | 84.22 ↑ 0.25 | 75.13 ↓ 0.77 | 2272.01 ↑ 29.31 |
| Ours | InternLM-XComposer2-VL-7B | 10k | 84.38 ↑ 0.41 | 75.45 ↓ 0.45 | 2276.30 ↑ 33.60 |
| Direct | InternLM-XComposer2-VL-7B | 32k | 84.57 ↑ 0.60 | 75.14 ↓ 0.76 | 2245.15 ↑ 2.450 |

As shown in the table, our algorithm improves the baseline algorithm (random sampling) by 0.50 on the UC Merced dataset and 0.67 on the AID dataset, with an average improvement of 0.58. In contrast, the KCenterGreedy clustering algorithm improves by 0.64 on the UC Merced dataset but decreases by 3.90 on the AID dataset, resulting in an overall decrease of 1.63 compared to the baseline algorithm. Overall, our algorithm achieves the best training performance.

To further observe the improvement of our algorithm over the baseline algorithm, we tested the training performance on a dataset of 10,000 entries and on the entire classification dataset. The results are shown in Table 3.

As shown in the table, when the dataset size is expanded to 10,000 entries, our algorithm shows even greater advantages, improving by 0.63 on the AID dataset and by 1.77 on the UC Merced dataset compared to the baseline algorithm, with an overall improvement of 1.20. The average improvement of 0.58 from 5000 to 10,000 entries is nearly double, indicating that the performance improvement brought by our algorithm increases with the dataset size. Additionally, when training on the entire 32k dataset, our algorithm, using only 10k entries, is only 1.42 points lower on the UC Merced dataset and 2.64 points lower on the AID dataset, with an overall average decrease of 2.00. This result demonstrates that our algorithm can significantly approximate the performance of training on the entire dataset with just one-third of the data.

Furthermore, we compared the performance of models trained with our algorithm and the baseline algorithm in general domains. The results are shown in Table 4.

As shown in the table, our algorithm also retains the best general domain capabilities, demonstrating superior performance over the random sampling method on the MMBench_DEV_en, SEEDBench, and MME datasets, achieving scores of 84.38, 75.45, and 2276.30, respectively. The performance on MMBench_DEV_en and SEEDBench exceeds that of the original model, with improvements of 0.41 and 33.60, respectively. In contrast, while direct training on the 32k dataset shows an improvement on MMBench_DEV_en, it slightly declines on SEEDBench. Overall, our method significantly enhances performance metrics in the remote sensing domain while maintaining the model's general capabilities, demonstrating its effectiveness and superiority.

*D. Optimal training data ratio*

To determine the optimal training data ratio, we conducted a detailed comparison of training durations and model performance for different data volumes (5000, 10000, 15000, and 32000 samples). The experimental results are shown in Figure 8.

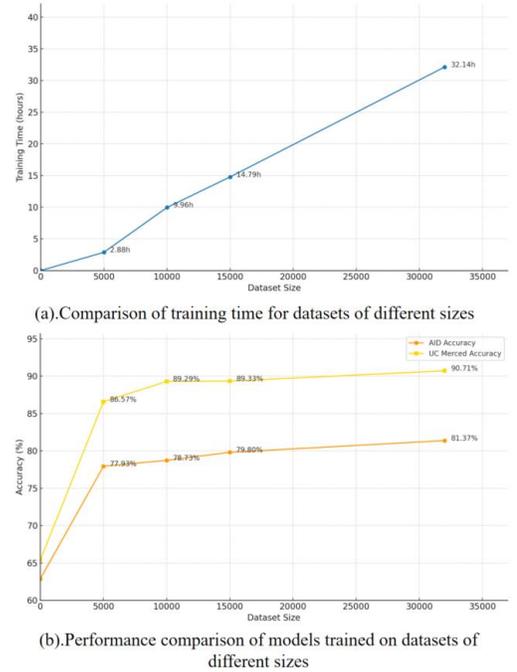

(a).Comparison of training time for datasets of different sizes

(b).Performance comparison of models trained on datasets of different sizes

**Fig. 8.** Comparison of training time and model performance under different sizes of datasets

As illustrated in Figure 8, increasing the training data volume leads to improved model performance on both the AID and UC Merced datasets. Specifically, with 5000 samples, the performance on the AID dataset is 77.93, and on the UC Merced dataset, it is 86.57. When the data volume is increased to 10000 samples, the performance on the AID and UC Merced datasets rises to 78.73 and 89.29, respectively. Further increasing the data volume to 15000 and 32000 samples results in performance levels of 79.80 and 81.37, as well as 89.33 and 90.71. This indicates that more data generally improves model performance, but the performance gain gradually diminishes.

The training duration data show a significant increase with the data volume. For instance, training with 5000 samples takes 2.88 hours, while training with 32000 samples increases to 32.14 hours, an additional 29.26 hours.

TABLE V
COMPARE THE EVALUATION RESULTS OF DIFFERENT MODELS ON AID AND UCMERCED DATASETS

| Model | AID | UCMerced | Avg. |
|---|---|---|---|
| MiniGPTv2 [38] | 4.76 | 12.90 | 8.83 |
| Qwen-VL-Chat [39] | 62.90 | 52.60 | 57.75 |
| LLaVA-1.5 [40] | 68.00 | 51.00 | 59.5 |
| InternLM-XComposer2-VL-7B | 62.87 | 65.38 | 64.13 |
| GeoChat | 72.03 | 84.43 | 78.23 |
| **Ours** | **77.19** | **89.86** | **83.53** |

TABLE VI
COMPARE THE EVALUATION RESULTS OF DIFFERENT MODELS ON THE LRBEN DATASET

| Model | RSVQA-LR | | | |
| | Rural/Urban | Presence | Compare | Avg. |
|---|---|---|---|---|
| LLaVA-1.5 | 59.22 | 73.16 | 65.19 | 65.86 |
| InternLM-XComposer2-VL-7B | 69.00 | 52.62 | 70.80 | 64.14 |
| MiniGPTv2 | 60.02 | 51.64 | 67.64 | 59.77 |
| InstructBLIP [41] | 62.62 | 48.83 | 63.92 | 59.12 |
| Mplug-Owl2 [42] | 57.99 | 74.04 | 65.04 | 65.69 |
| Qwen-VL-Chat | 62.00 | 47.65 | 54.64 | 58.73 |
| SkyEyeGPT [43] | 88.93 | 88.63 | 75.00 | 84.16 |
| RSGPT | 94.00 | 91.17 | 91.70 | 92.29 |
| GeoChat | 91.09 | 90.33 | 94.00 | 91.81 |
| LHRS-Bot [44] | 89.07 | 88.51 | 90.00 | 89.19 |
| **Ours** | **89.00** | **91.91** | **91.78** | **90.90** |

By comparing model performance and training durations across different data volumes, we found that with 10000 samples, the model's performance is close to its peak, while the training duration is significantly lower compared to 15000 and 32000 samples. Specifically, the performance difference between 10000 and 32000 samples is an average of 2.13, with a reduction in computation cost by 22.18 hours.

In summary, with 10000 samples, the model achieves a high performance while significantly reducing training time and computational resources. Thus, 10000 samples represent the optimal balance between performance and computational cost. This indicates that using approximately 1/3 of the total dataset achieves better training results while substantially lowering the computational cost.

*E. Final Performance of Our Algorithm*

Using our algorithm for automatic clustering, we divided the RS multimodal instruction following dataset into 7 categories, as shown in the vector space visualization in Figure 9.

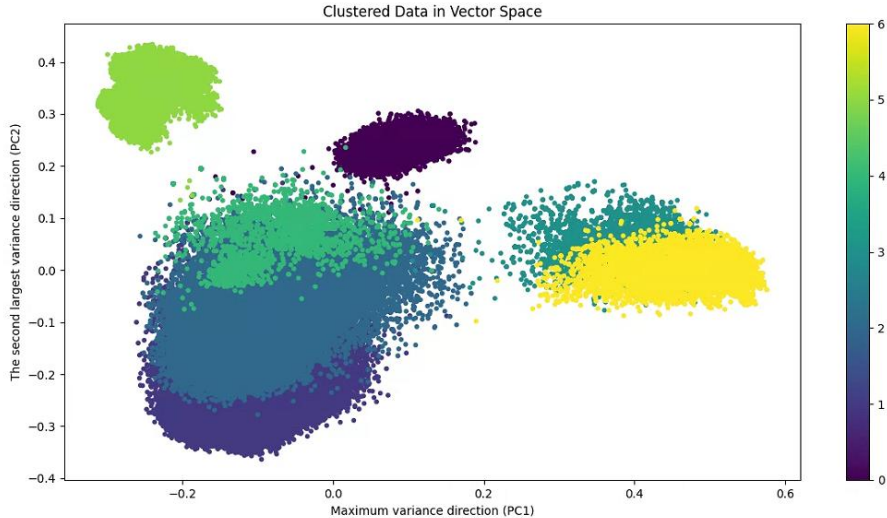

**Fig. 9.** RS dataset clustering in vector space.

We then selected 15,000 data entries from each category, totaling 105,000 entries for training. The model was trained for three epochs, and the results are shown in Tables 5 and 6.

As shown in the tables, the model trained with only 105k entries achieved 77.19 on the AID dataset and 89.86 on the UC Merced dataset, which are 5.16 and 5.43 points higher than GeoChat, respectively. On the LRBEN dataset, it achieved an average of 90.90, only 0.91 points lower than GeoChat. Observing the performance of the original models on the AID, UC Merced, and LRBEN datasets, we find that our original model InternLM-XComposer2-VL-7B outperforms GeoChat's original model LLaVA-1.5 by an average of 4.63 on AID and UC Merced. After training, our model outperforms GeoChat by 5.3 on these datasets. On the LRBEN dataset, InternLM-XComposer2-VL-7B scores 1.72 points lower than LLaVA-1.5, and our final trained model scores 0.91 points lower than GeoChat.

These results indicate that the performance of the original model has a direct positive impact on the final training performance. However, the key finding is that by selecting high-quality, generalizable datasets, our algorithm can achieve results comparable to those obtained from training on the full dataset, using only one-third of the data. This demonstrates the effectiveness and efficiency of our method in enhancing model performance.

*F. Ablation Study*

To further evaluate the performance of our algorithm, we compared the results of training on the entire dataset versus a 105k subset selected by our algorithm, both using InternLM-XComposer2-VL-7B on two 3090 GPUs for one epoch. The results are shown in Tables 7, 8, and 9. Notably, training on the 105k dataset took approximately 35 hours, while training on the full 318k dataset required around 110 hours, more than three times the time consumption.

TABLE VII
COMPARE THE EVALUATION RESULTS OF MODELS TRAINED ON DATA SETS OF DIFFERENT SCALES ON AID AND UC MERCED

| Method | Size | AID | UC Merced | Avg. |
|---|---|---|---|---|
| Ours | 105k | 75.60 | 85.67 | 80.64 |
| Direct | 318k | 75.07 ↓ 0.53 | 87.95 ↑ 2.28 | 81.51 ↑ 0.87 |

TABLE VIII
COMPARE THE EVALUATION EFFECTS OF MODELS TRAINED ON DATA SETS OF DIFFERENT SCALES ON LRBEN

| Method | RSVQA-LR | | | |
|---|---|---|---|---|
|  | Rural/Urban | Presence | Compare | Avg. |
| Ours | 90.00 | 90.73 | 91.05 | 90.59 |
| Direct | 92.00 ↑ 2.00 | 91.57 ↑ 0.84 | 92.45 ↑ 1.40 | 92.01 ↑ 1.42 |

TABLE IX
COMPARE THE EVALUATION EFFECTS OF MODELS TRAINED ON DATA SETS OF DIFFERENT SCALES IN GENERAL FIELDS

| Method | Model | Size | MMBench | Seedbench | MME |
|---|---|---|---|---|---|
| / | InternLM-XComposer2-VL-7B | / | 83.97 | 75.9 | 2242.70 |
| Ours | InternLM-XComposer2-VL-7B | 105k | 83.78 ↓ 0.19 | 74.92 ↓ 0.98 | 2121.01 ↓ 121.69 |
| Direct | InternLM-XComposer2-VL-7B | 318k | 83.75 ↓ 0.22 | 74.18 ↓ 1.72 | 1982.90 ↓ 259.80 |

As seen in Tables 7 and 8, the performance difference between training on the entire dataset and the 1/3 subset selected by our algorithm is minimal in remote sensing tasks. On the AID dataset, our algorithm even achieved an accuracy that is 0.53% higher than training on the full dataset. Our algorithm reached an accuracy of 80.64 on the AID and UC Merced evaluation datasets, which is only 0.87% lower than training on the full dataset. On the RSVQA-LR dataset, our algorithm averaged an accuracy of 80.59, just 1.42% lower than the full dataset training.

It is worth noting that the training results on the UC Merced and AID datasets are not as high as those achieved by training on a single type of dataset as described in Section 4.3. This indicates that training on datasets of different types together can lead to significant data conflicts. However, our method achieves a higher score on the AID dataset compared to training on the entire dataset, suggesting that selecting high-quality subsets can alleviate some of the data conflicts.

It's worth noting that in general-domain tasks, our algorithm retained more performance than training directly on the full dataset, achieving scores of 83.78, 74.92, and 2121.01 on MMBench, Seedbench, and MME, respectively—all higher than the performance scores of the model trained on the full dataset. Additionally, on the Seedbench and MME datasets, the accuracy loss from training on the full dataset was nearly twice that of the loss from our algorithm.

In summary, our algorithm saves more than twice the training time while maximizing the retention of general-domain capabilities, with only about a 1% accuracy loss in the remote sensing domain.

## V. CONCLUSION

This study addresses the issue of data selection for multimodal large models in various domain tasks by proposing an adaptive fine-tuning algorithm. Most current research directly trains on large-scale multimodal data, which not only requires substantial computational resources but also results in significant performance degradation when randomly selecting a small subset of data. To resolve this, we first project the large-scale data into vector space and use the MiniBatchKMeans algorithm for automated clustering. Then, we measure the generalizability of the data by calculating the translation difference in the multimodal large model's vector space between the original and perturbed data, and autonomously select data with high generalizability for training.

Our experiments, based on the InternLM-XComposer2-VL-7B model, were conducted on the remote sensing multimodal dataset proposed by GeoChat. The results show that using the adaptive fine-tuning algorithm, our method outperforms the random sampling and KCenterGreedy clustering algorithms in training with a 5,000-entry dataset, achieving the best domain and general performance with a 10,000-entry dataset. Ultimately, using only 105,000 data entries—one-third of the GeoChat dataset—and training on a single 3090 GPU, our model achieved performances of 89.86 on the UC Merced dataset and 77.19 on the AID dataset, which are 5.43 and 5.16 points higher than GeoChat, respectively. On the LRBEN evaluation dataset, our model was only 0.91 points lower on average. Furthermore, comparing the performance of models trained on the full dataset versus our one-third dataset, we found that our approach reduced training time by more than 68.2% while maintaining general-domain capabilities with only a 1% average decrease in remote sensing accuracy.

In summary, our adaptive fine-tuning algorithm effectively selects high-quality data, enhancing model performance in specific domains while maintaining general performance under limited computational resources. This algorithm has significant practical value for training multimodal large models, especially in scenarios with constrained computational resources.